\documentclass[letterpaper, 10pt, conference]{ieeeconf}
\IEEEoverridecommandlockouts
\usepackage{cite}
\usepackage{amsmath,amssymb,amsfonts}
\usepackage{algorithmic}
\usepackage{graphicx}
\usepackage{textcomp}
\usepackage{xcolor}
\usepackage{subcaption}
\usepackage[utf8]{inputenc}
\usepackage[export]{adjustbox}
\usepackage{wrapfig}
\def\BibTeX{{\rm B\kern-.05em{\sc i\kern-.025em b}\kern-.08em
    T\kern-.1667em\lower.7ex\hbox{E}\kern-.125emX}}
\begin{document}

\title{\LARGE \bf
Navigation with Tactile Sensor for Natural Human-Robot Interaction
}

\author{Zhen Hao Gan$^{1,2}$, Yangwei You$^{2}$, Meng Yee (Michael) Chuah$^{2}$
%, Jiacheng Andrew Leong$^{2}$
%, Tai Pang Chen$^{2}$
\thanks {$^{1}$Electrical and Electronic Engineering, Nanyang Technological University, 639798, Singapore, {\tt\small ganz0015@e.ntu.edu.sg}}
\thanks {$^{2}$Institute for Infocomm Research, Agency for Science, Technology and Research, 138632, Singapore, {\tt\small youy@i2r.a-star.edu.sg}}
}

\maketitle

\begin{abstract}
Tactile sensors have been introduced to a wide range of robotic tasks such as robot manipulation to mimic the sense of human touch. However, there has only been a few works that integrate tactile sensing into robot navigation. This paper describes a navigation system which allows robots to operate in crowded human-dense environments and behave with socially acceptable reactions by utilizing semantic and force information collected by embedded tactile sensors, RGB-D camera and LiDAR.
Compliance control is implemented based on artificial potential fields considering not only laser scan but also force reading from tactile sensors which promises a fast and reliable response to any possible collision.
In contrast to cameras, LiDAR and other non-contact  sensors, tactile sensors can directly interact with humans and can be used to accept social cues akin to natural human behavior under the same situation. 
Furthermore, leveraging semantic segmentation from vision module, the robot is able to identify and, therefore assign varying social cost to different groups of humans enabling for socially conscious path planning. 
At the end of this paper, the proposed control strategy was validated successfully by testing several scenarios on an omni-directional robot in real world.
\end{abstract}

\section{Introduction}

\begin{figure}[t]
\centerline{\includegraphics[width=\linewidth]{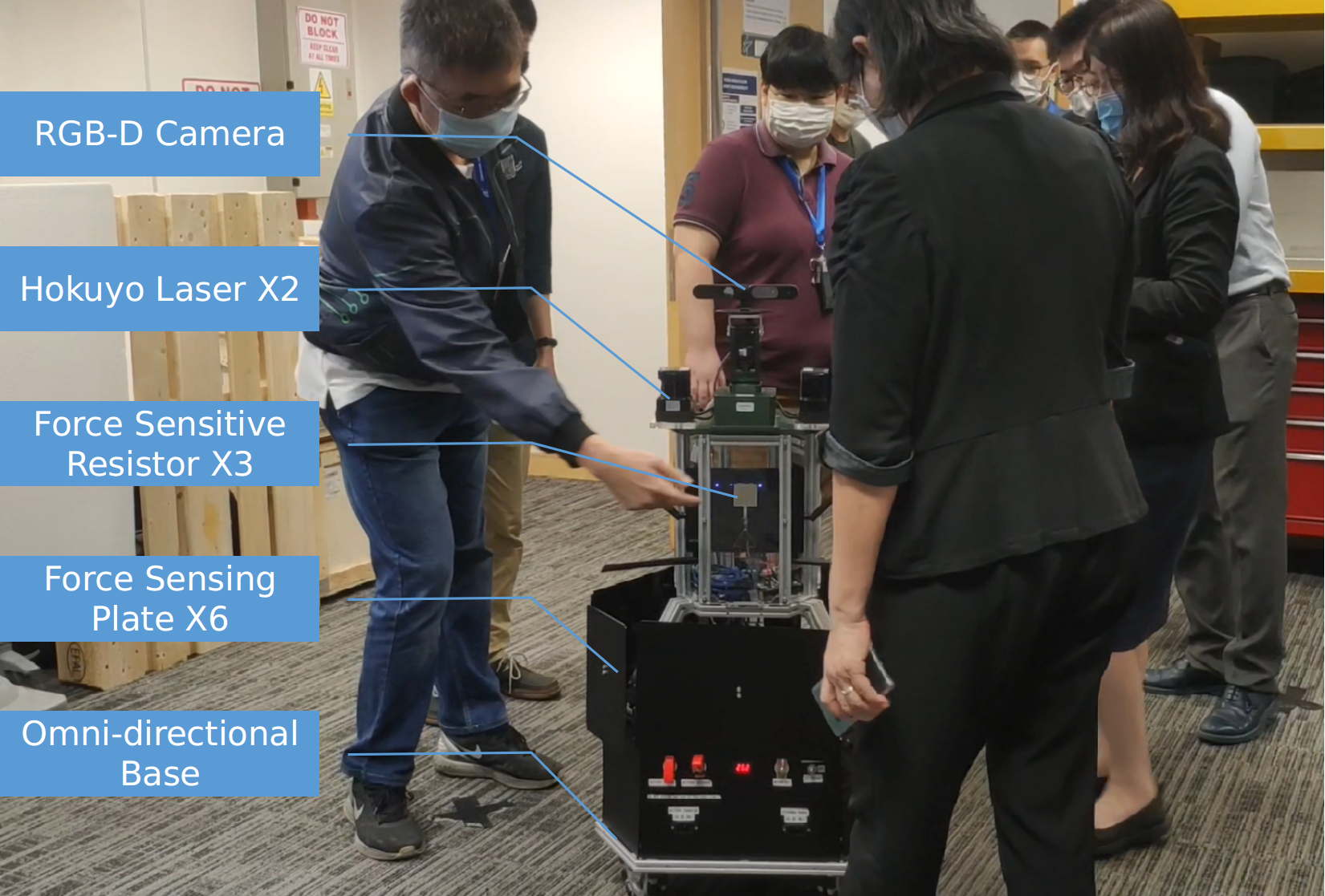}}
\caption{An Omni-directional robot moving towards its goal in human-rich environment. The robot is able to adapt to human-robot interaction through camera, LiDAR and tactile sensing inputs for safe navigation.}
\label{fig:omni}
\end{figure}

In the recent years, sensing technologies and computing capabilities has been rapidly growing and advancing which sparks great interest in robotic navigation. Specifically, the application of automated ground vehicles in complex human-rich environments has been explored widely using different types of sensors and navigation algorithm. The robots in this application are expected to reach goals as fast as possible yet navigate safety around humans in a dynamic environment as shown in Fig. \ref{fig:omni}. The problem is classified as challenging due to the unpredictability and generally unknown intent of each individual in the environment. Furthermore, the robot must comply with the social interactions using the limited range of view of on board sensors\cite{b1}. 

Recent approaches to navigate through a dense-crowded environment are mostly divided to model-based and learning-based. Learning-based methods focus on developing a policy that can learn to react based on the current observations about the environment, among which the most popular one is reinforcement learning \cite{b2}.
One example is that reinforcement learning can understand human-human or human-robot interactions and react accordingly after training \cite{b3, b4}.
Inverse Reinforcement Learning \cite{b5} also has been explored to understand the social cost from human expert \cite{b6}. Although learning-based methods can achieve near-human reaction and trajectory, they tend to suffer from high computing cost from either during training or predicting human intentions and social interactions between humans and robots. For an unknown environment or unknown obstacles, these algorithms may fail as they are not accounted for during training. These limitations can be a high entry cost for general utilization of such robots.

On the other hand, model-based navigation methods attempt to model the human-human and human-robot interactions, and incorporate the information into multi-agent collision avoidance algorithms. These methods normally divide the prediction for social interactions and the path planning into two different steps. The most basic approach would rely on a reactive planner to do one-step look-ahead path planning and treat human as a simple, static obstacle \cite{b7} apply some other social force models \cite{b8, b9}. 
This might generate unsafe planned trajectories or movements that could potentially result in human-robot collision because an accurate and precise description of every possible interaction is not always possible and this does not account for possible human intent to move around in a dynamic environment. 
Several research works in trajectory prediction have been proposed to model the human intention in the environment as a predicted path and incorporate it into path planning algorithms. However, this could still cause collision with humans if the prediction is too noisy or unreliable. Another problem can arise from the navigation system is the freezing robot problem \cite{b10} where the planner assumes that a human is impermeable and does not react to an approaching robot, this is not case for normal human behaviour. This will result unnatural interaction between human and robot as the robot will always plan to avoid any possible collision with humans and may result in the robot "freezing" if there is no free, clear space for the robot to plan a path.

In this work, it is proposed to introduce human-like social behaviour by introducing an array of tactile sensors around the robot and utilizing model-based navigation methods to  navigate in human-rich environments and mitigate the effects of possible human-robot collisions. Tactile sensing enables the robot to perceive and mitigate the severity of human-robot collisions arising from unpredictable human motion during a planned motion. Our proposed method is low-cost, easy to implement, and transferable to other mobile robots. 

\section{Background}

\subsection{Tactile Sensors}
Exteroceptive sensors such as vision-based sensors and laser-based sensors are commonly used in robotic applications to perceive the environment. Tactile sensors which typically consist of passive sensors with the capability to identify some physical properties of an entity in contact with the sensor apparatus, are rarely used for perceiving the environment in the context of navigation. These sensors comprise many different operational types; traditional strain gauges, electromagnetic device sensors, force-sensitive resistors, capacitive tactile array, optical devices, piezoelectric resonance, and shape-memory alloy devices as micro coil actuators used for 2-D and 3-D tactile displays, etc \cite{b11, b12, b13}. Tactile senses are useful for robot operation due to their ability to mimic the sense of touch, allowing the robot platform to be able to process textural or contact information, which can then be used to understand objects or the environment. There has been an increasing popularity of integrating tactile sensors for robot manipulation tasks but not so for the purpose of robot navigation.

\subsection{Layered Cost Map}

The fundamental methodology of any model-based navigation method is a reactive planner which utilize some cost function applied to an occupancy grid to produce a cost map of the environment for planning. Different types of cost map layers have been developed for various purposes and eventually combined together to generate a complete, useful cost map for path planning. In the context of navigation involving human-robot interaction, the cost map can be generally classified and divided into 4 distinct layers as follows; i) Static Layer, ii) Obstacle Layer, iii) Inflation Layer, iv) Proxemic Layer. 
The cost values over all the aforementioned layers will be considered following a specific rule (overwriting, taking maximum etc.) and form a final composite cost map for path planning \cite{b14}.

The Static Layer typically contains non-movable object in the environment such as walls and cupboards and can be generated by SLAM algorithm. The occupancy grid generated from this static map can then transformed into the first layer of cost map, setting the highest possible cost to all occupied grid cells making these cells impermeable by the planner. 

The second layer, Obstacle Layer cost map is generated via the input from sensors such as LiDAR and RGB-D camera. The extra obstacle detected in this layer will normally be reflected as an occupied, non-permeable space in the cost map. Normally the obstacles will be set to the highest cost in the cost map, but there exists application to assign dynamic cost on navigating in different terrains \cite{b15}

The third layer, Inflation Layer cost map is used to create an extra buffer area around the obstacles identified in the previous layer of cost map. This layer considers the approximate size of the robot and inflates a region equal to this size around the obstacle with a cost value which is one less than the highest cost. This process creates a non-permeable bubble that the path planners are not allowed plan through. The remaining inflation radius will record an exponentially decreasing cost tending to zero in order to create a bubble that discourages the planner from planning a path close to the obstacle. 

Finally, the last layer, Proxemic Layer is an extension from the Inflation Layer to capture the human-robot interaction. The human intent in an environment such as the heading and velocity are be predicted by computer vision algorithm. Using these predictions of human intent, the cost map can further modelled a detected, static person into a moving person through some Gaussian-distributions. This process basically extended the cost of moving person in the direction of intended moving, thus increase the cost of robot to pass through the path of moving person. 

All these layers combined to generate a simple global and local cost map that the reactive path planner can utilize to generate a path for the robot to transverse in the environment. This method is the backbone of ROS basic navigation stack and heavily relies on both the accurate and precise prediction of human intent and sensors input which can caused collision from time to time. In our work, we intended to overcome this problem with tactile sensors on board to form a proximity filter and utilize a computer vision algorithm to locate human for dynamic cost assignation. The robot with our solution can engage actively with human to increase social interaction and prevent high cost of collision using tactile sensors.

\subsection{Contribution}
Compared with existing works on social navigation of mobile robots, our contributions are mainly threefold:
\begin{enumerate}
\item{Based on artificial potential field, compliance control is improved by combining force reading and laser scan to quickly response to environment changing in both zero and positive proximity.}
\item{The interaction capability of tactile sensors is explored to deliver social cues akin to human behaviors for more natural and efficient navigation in human-rich scenarios.}
\item{Given semantic segmentation from robot vision, specific social costs are assigned to different groups of people allowing for socially conscious path planning.}
\end{enumerate}

This paper is organized as following: Section \ref{sec:approach} introduces the technical details of how the compliance control and social navigation strategy are implemented. Thereafter, the proposed methods are validated in Section \ref{sec:experiment} by carrying out several testing scenarios on an omni-directional robot in real world. In the end, the paper summarizes in Section \ref{sec:conclusion} and gives an outlook on future research.

\begin{figure}[t]
\centerline{\includegraphics[width=5cm, angle=90]{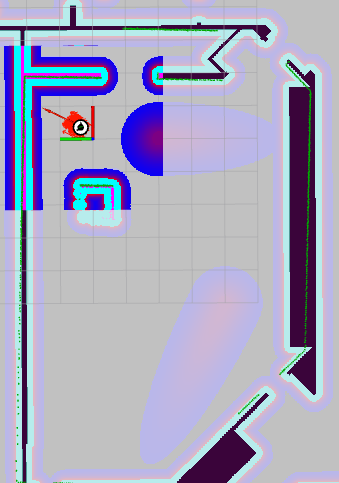}}
\caption{Example of Semantic Layer working together. Our approach can assign dynamic cost to human and capture the human intent in the environment.}
\label{fig:semantic_layer}
\end{figure}

\section{Approach}
\label{sec:approach}
While most common perception sensors like camera and LiDAR have been massively deployed on robots, it can still miss some details due to blind spot. The basic approach to construct a useful navigation stack could cause collision with moving obstacles such as pedestrians at some point due to unreliable prediction of human intention and obstacles. The following section will describe the method to integrate dynamic cost for human and tactile sensor to encourage social interaction between human and robots. We first describe how the tactile sensors and proximity filter are deployed to our robot and second part would layout the details of semantic layers which integrate dynamic cost for human using the tactile sensors and semantic input of computer vision algorithm.

\subsection{Tactile Sensor and Proximity Filter}
Any impact acting on the robot can be detected by force tactile sensors and passed to a compliance controller, named proximity filter here. The proximity filter considers both the tactile response and the laser scan in a dynamic situation while ignores laser input when the robot is static.
With tactile sensors, the robot is able to detect very near obstacles which possibly lie in the blind spot of laser and responses accordingly.
The velocity command given by a local planner will be temporarily adjusted until the proximity filter exceeded its timeout.
This maneuver efficiently avoids severe collision and running further into any surrounding obstacles. 
In addition, the repulsive maneuver can trigger a re-planning from the local path planner and relieve from being stuck in local minima. Both LiDAR, tactile sensor and proximity filter can be integrated together to minimise the maintenance of severe collision. This forms the passive social interactions with human which mainly focuses on avoiding more severe collision. 

Other than avoiding collision, the robot can utilize the tactile sensor to engage actively with human in the near vicinity and increase its social interaction. For example, the robot can response to any contact by human when it is in static mode. The robot will then turn to the direction of human, identify the human and assign a cost to it. This could be useful in a hospital environment to identify a child when the child is lower than the height of LiDAR sensors and avoid collision with the child.
It can also be helpful for human that actively require the robot to assign special cost during path planning. This is meaningful and allows for more human-robot interaction.

\begin{figure}[t]
\begin{subfigure}{0.3\textwidth}
\centerline{\includegraphics[height=5cm,]{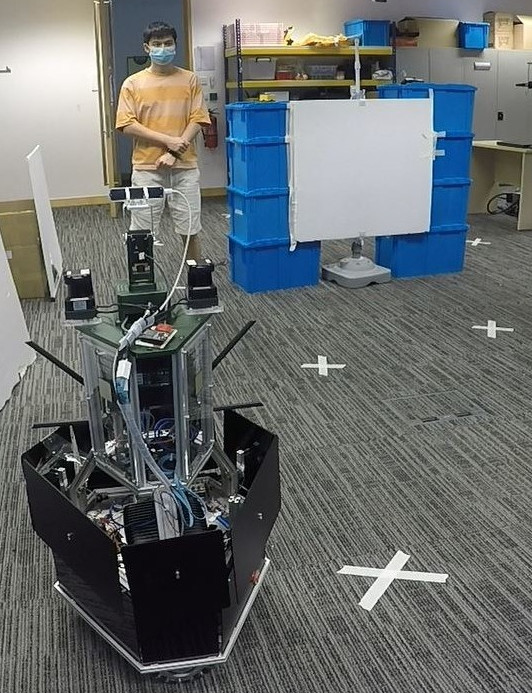} }
\end{subfigure}
\begin{subfigure}{0.1\textwidth}
\centerline{\includegraphics[height=5cm,]{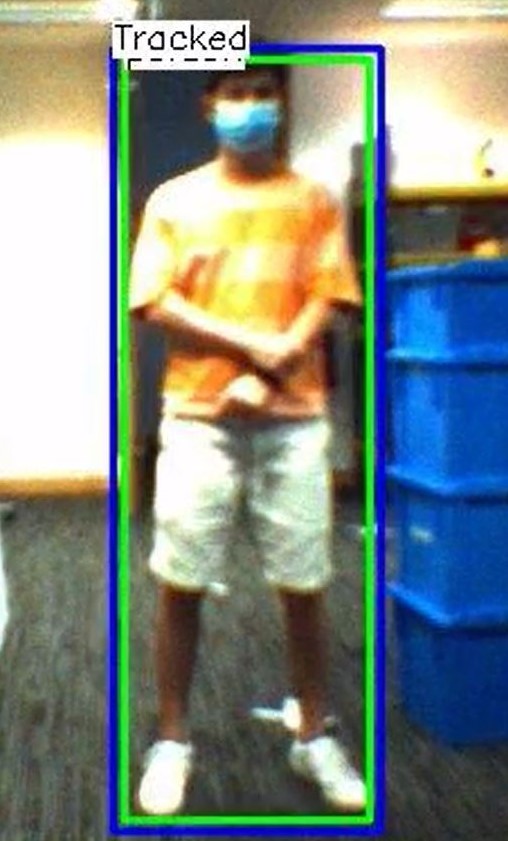} }
\end{subfigure}
\caption{Experiment setup and detected human. Left: two exits are presented to the robot (two ends of the blue containers) and the goal is directly behind the human; Right: detected human marked by a bounding box through semantic information from RGB-D camera.}
\label{fig:experiment_setup}
\end{figure}

\begin{figure}[t]
\centering
\begin{subfigure}{0.23\textwidth}
\centerline{\includegraphics[height=5cm,width=0.95\linewidth]{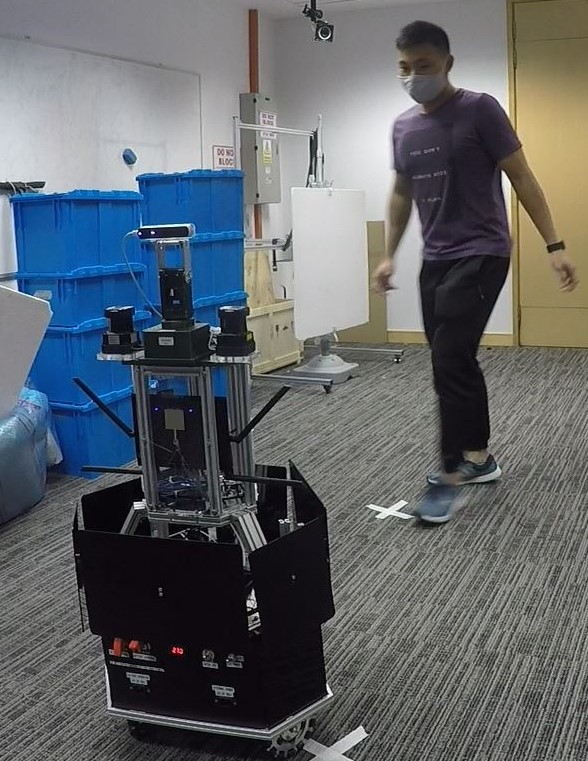}}
\caption{}
\label{fig:reaction1}
\end{subfigure}
\begin{subfigure}{0.23\textwidth}
\centerline{\includegraphics[height=5cm, width=0.95\linewidth]{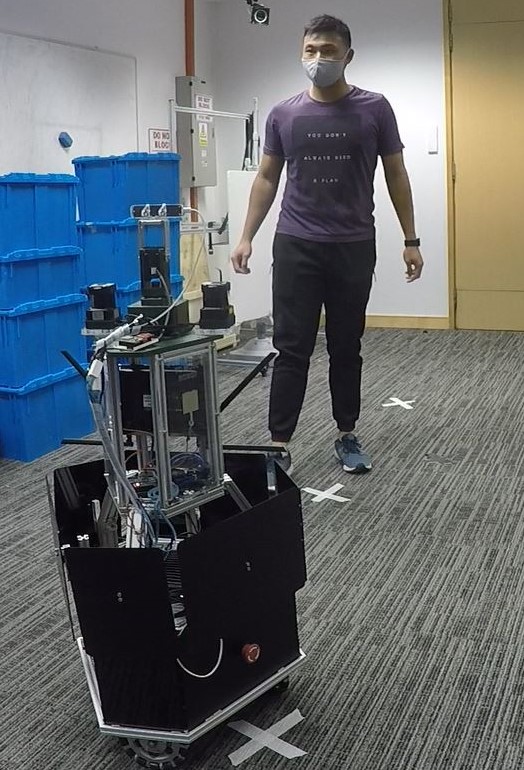} }
\caption{}
\label{fig:reaction2}
\end{subfigure}

\begin{subfigure}{0.23\textwidth}
\centerline{\includegraphics[height=4cm,width=0.95\linewidth]{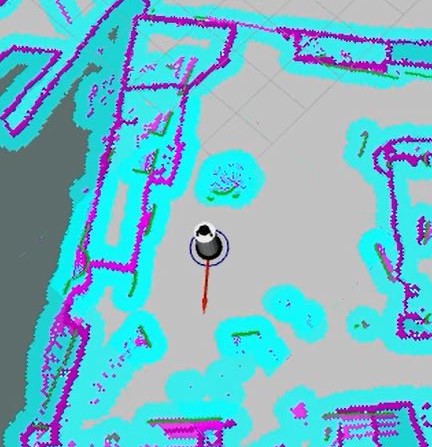} }
\caption{}
\label{fig:reaction3}
\end{subfigure}
\begin{subfigure}{0.23\textwidth}
\centerline{\includegraphics[height=4cm,width=0.95\linewidth]{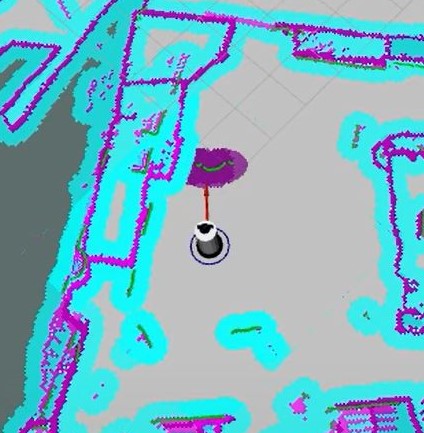} }
\caption{}
\label{fig:reaction4}
\end{subfigure}
\caption{Reacting to human contact.  (a), (b) are the experiment snapshots and (c), (d) are the corresponding cost map. (a) Human approach to touch robot from behind. (b) The robot turn towards to the direction of human. (c) Unidentified obstacle, highest cost (cyan) assigned. (d) Identified obstacle as human after rotation, lower cost (purple) assigned}
\label{fig: reaction}
\end{figure}

\begin{figure*}[ht]
\centering
\begin{subfigure}{0.18\textwidth}
\centerline{\includegraphics[width=0.95\linewidth, height=4cm]{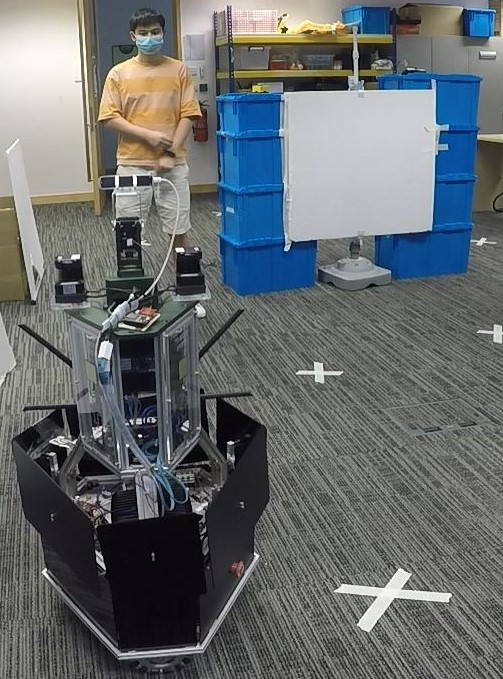} }
\caption{}
\label{fig:blocked path1}
\end{subfigure}
\begin{subfigure}{0.18\textwidth}
\centerline{\includegraphics[width=0.95\linewidth, height=4cm]{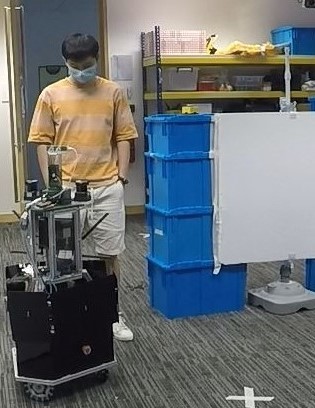}}
\caption{}
\label{fig:blocked path2}
\end{subfigure}
\begin{subfigure}{0.18\textwidth}
\centerline{\includegraphics[width=0.95\linewidth, height=4cm]{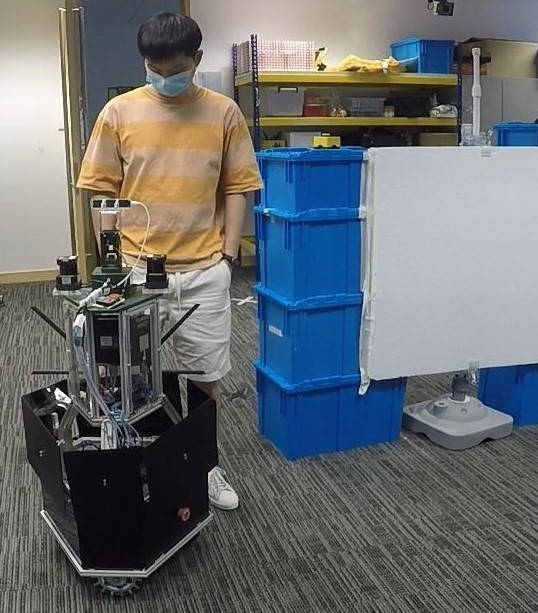}}
\caption{}
\label{fig:blocked path3}
\end{subfigure}
\begin{subfigure}{0.18\textwidth}
\centerline{\includegraphics[width=0.95\linewidth, height=4cm]{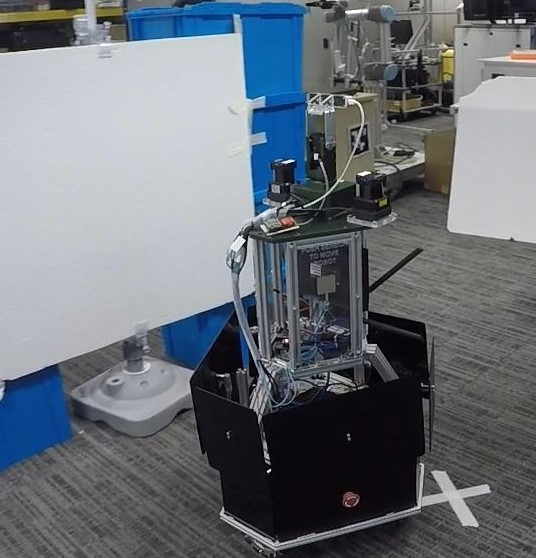}}
\caption{}
\label{fig:blocked path4}
\end{subfigure}
\begin{subfigure}{0.18\textwidth}
\centerline{\includegraphics[width=0.95\linewidth, height=4cm]{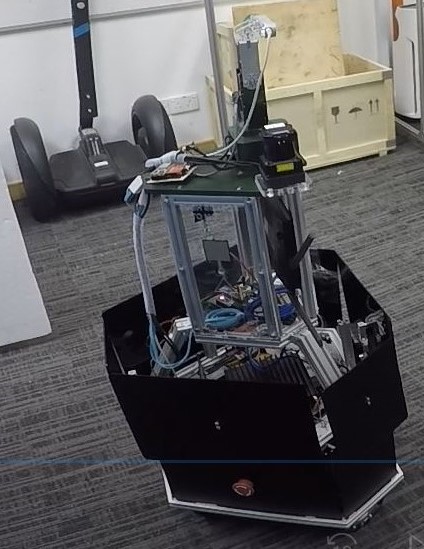}}
\caption{}
\label{fig:blocked path5}
\end{subfigure}

\begin{subfigure}{0.18\textwidth}
\centerline{\includegraphics[trim=0 0.3cm 0 0, clip, width=0.95\linewidth, height=3cm]{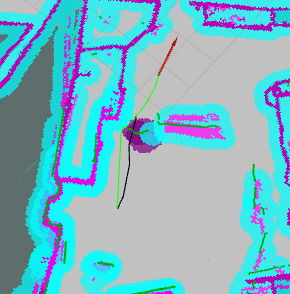} }
\caption{}
\label{fig:blocked path6}
\end{subfigure}
\begin{subfigure}{0.18\textwidth}
\centerline{\includegraphics[trim=0 0.3cm 0 0, clip, width=0.95\linewidth, height=3cm]{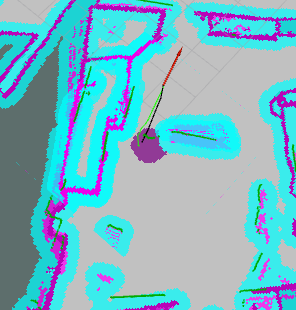}}
\caption{}
\label{fig:blocked path7}
\end{subfigure}
\begin{subfigure}{0.18\textwidth}
\centerline{\includegraphics[trim=0 0.3cm 0 0, clip, width=0.95\linewidth, height=3cm]{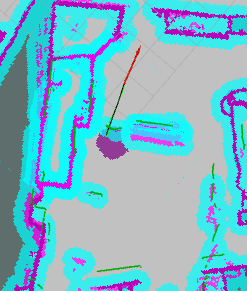}}
\caption{}
\label{fig:blocked path8}
\end{subfigure}
\begin{subfigure}{0.18\textwidth}
\centerline{\includegraphics[trim=0 0.3cm 0 0, clip, width=0.95\linewidth, height=3cm]{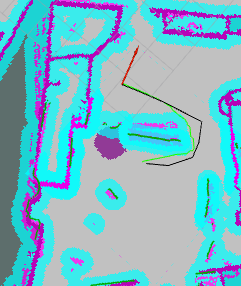}}
\caption{}
\label{fig:blocked path9}
\end{subfigure}
\begin{subfigure}{0.18\textwidth}
\centerline{\includegraphics[trim=0 0.3cm 0 0, clip, width=0.95\linewidth, height=3cm]{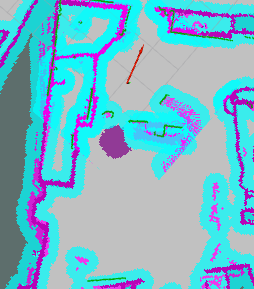}}
\caption{}
\label{fig:blocked path10}
\end{subfigure}
\caption{Human blocked way to the robot after touching. (a)-(e) are the experiment snapshots and (f)-(j) are the corresponding cost map. (a) Initialization of experiment. (b) Robot approach the goal via nearest exit. (c) Human blocked path after contact. (d) Re-plan path to further exit. (e) Goal reached. (f) Path planned to goal. (g) Lower cost (purple) for human before contact. (h) Highest cost (cyan) assigned after blocked path confirmed. (i) Path planner re-plan to further exit. (j) Goal reached.}
\label{fig:blocked path}
\end{figure*}

\subsection{Semantic Layer}
When human is gently touched by surrounding pedestrians in crowded situations, we usually subconsciously move aside to give ways to others.
This behavior is similar to what we try to develop for robots in previous proximity filter.
However, in some extremely crowded cases, human may find that people around are not willing or impossible to give way. Human can easily handle this situation by looking for alternative exits or waiting for the blocked exit to be free.This scenario leads us to think we can achieve such navigation maneuver through dynamic human assignation which the usual ROS navigation stack cannot fulfill. 

A Semantic Layer cost map has been introduced to mimic this human behaviour. This cost map is constructed based on the same four usual layers of cost map mentioned above but with some modification to the first three layers. The modified version includes: i) Static Layer, ii) Semantic Obstacle Layer, iii) Semantic Inflation Layer, iv) Semantic Proxemic Layer. The modifications are made to adapt the inflation and model the human intent from the Semantic Obstacle Layer while the original layered cost map only considers the highest cost for each map cell.

In the Semantic Obstacle Layer cost map, we assume every person segmented out by the vision module has a more permeable cost while the Semantic Inflation and Semantic Proxemic Layer inflates from the corresponding cost using information such as approximate robot radius, position, heading and velocity of human obstacles. 
In other words, human around the robot are always assumed as 'moveable gentleman' and willing to give ways.
This strategy can alleviate the 'freezing robot' problem in crowded scenarios where the robot always cannot find a valid path. Instead, the robot will try to squeeze into crowd and the safety can be still be guaranteed by the aforementioned proximity filter.

If the human blocking the shortest path is unwilling to move even after being gently 'touched' by the robot,
this will trigger the robot to move in opposite contact force direction and wait for timeout, acting as a safety measure to prevent further damage to the robot. 
In the meantime, the robot will observe if the human has unblocked the path and remain the permeable cost of human. 
If the human still persists to block the exit after timeout, the Semantic Obstacle layer will mark the place taking by this particular human as occupied and re-plan the path. 
In addition, the Semantic Inflation and Semantic Proxemic Layer will inflate the cost assigned in Semantic Obstacle Layer accordingly based on the robot inscribed radius and human intention to move as shown in Fig. \ref{fig:semantic_layer}.

The Semantic Layer enables a more active social interaction between robots and human by using tactile sensing and semantic information from vision and mainly focuses on assigning permeable cost and engaging actively with human to find an optimal path.
This approach can be potentially deployed in human-crowded environments like hospital to avoid high cost of collision while assigning dynamic cost to different groups of people. For instance, vulnerable people, such as patients and pregnant women, should have a higher cost and vice versa.  
To be noted, the proposed control strategy is also computation light-weight. It can run on an embedded system in real-time.

\section{Experiment}
\label{sec:experiment}
The validation platform we used here was an omni-directional robot called Omnirobot as shown in Fig. \ref{fig:omni}. The tactile sensors that capture the force response has been installed around the lower level of robot as 6 force-sensing plates. The robot is also equipped with one Structure Core RGB-D camera and two LiDAR sensors at a higher level of the robot. The camera is used for its semantic information to predict the human location as shown in Fig. \ref{fig:experiment_setup}, while the LiDAR sensor can used to filter the precise human location, detect other non-movable obstacles and provide laser range information to the proximit filter. The localization of the robot is done through AMCL method \cite{b16} and the local path planner used is EBand Local Planner \cite{b17}. All the human detection, the construction of cost map and the computation of path planning algorithm are able to run real-time on a Intel IPC3 computer onboard of the robot.

\subsection{Passive Social Interactions - Navigating through crowded environment}
The first scenario is generated with human-crowded environment and constantly moving in the close vicinity of the robot. Our proposed proximity filter was built by utilizing semantic and force information collected by embedded tactile sensors, RGB-D camera and LiDAR. A compliance control based on the combination of artificial potential fields, force sensing input from tactile sensors and laser range information from LiDAR has been produced by the filter to generate reasonable and predictable controlled robot response to all combinations of inputs. With the proximity filter, the robot successfully navigated itself to the goal even with the planned path blocked and avoided several near-crash situation. This experiment demonstrated the capability of LiDAR, tactile sensors and proximity filter to perform passive social interactions with human and allow robots to operate in crowded human dense environments.

\subsection{Passive Social Interactions - Reacting to human contact}
The second scenario is generated as a human approached the robot vehicle and contact for attention in Fig. \ref{fig: reaction}. Initially, the human can not be identified by the camera due to limited field of view as shown in Fig. \ref{fig:reaction1}, thus the human was assigned to the highest cost possible in Fig. \ref{fig:reaction3}. After human contact, the robot sense a tactile response, turn to the direction of contact force (human) in Fig. \ref{fig:reaction2} and the camera identified the human in front of it using some computer vision algorithm. Although the human was moving around the in front of the robot, it was still able to track and detect the human. The human was detected and assign a more-permeable cost to human in Fig. \ref{fig:reaction4}, allowing more active social interactions with the human. This could be deploy in important environment such as hospital, where only the non-essential personnel can interact with robot while others interactions should be restricted, using the dynamic cost assignation to each human.

\subsection{Active Social Interactions - Path blocked after contact \label{ASI}}
In addition to the passive social interactions scenario we created for the robot, we have also setup an environment with two exits, one nearer to the goal (left) and one further away from goal (right) as shown in Fig. \ref{fig:experiment_setup} to experiment the active social interactions of the robot with human in Fig. \ref{fig:blocked path}. The goal was set directly behind the exit on the left with human blocking it in Fig. \ref{fig:blocked path1}, \ref{fig:blocked path6}. The robot first attempted to move towards the goal via the nearer exit as the total cost is lower to consider the further exit, but realized the exit is blocked by a human in Fig. \ref{fig:blocked path2}, \ref{fig:blocked path7}. The robot then attempted to engage with the human blockage through gentle touch and the proximity filter was kicked into work, repulsed and waited for the timeout in Fig. \ref{fig:blocked path3}. The human cost remained low and permeable before the timeout in Fig. \ref{fig:blocked path8}. In this scenario, the human decided to block the exit in Fig. \ref{fig:blocked path4}, thus the robot assigned the human to a higher cost as an indication of non-permeable obstacles in Fig. \ref{fig:blocked path9} and proceeded to re-plan its path to goal via the further goal as shown in Fig. \ref{fig:blocked path5}, \ref{fig:blocked path10}.

\subsection{Active Social Interactions - Path unblocked after contact}
\begin{figure*}[ht]
\centering
\begin{subfigure}{0.18\textwidth}
\centerline{\includegraphics[width=0.9\linewidth, height=4cm]{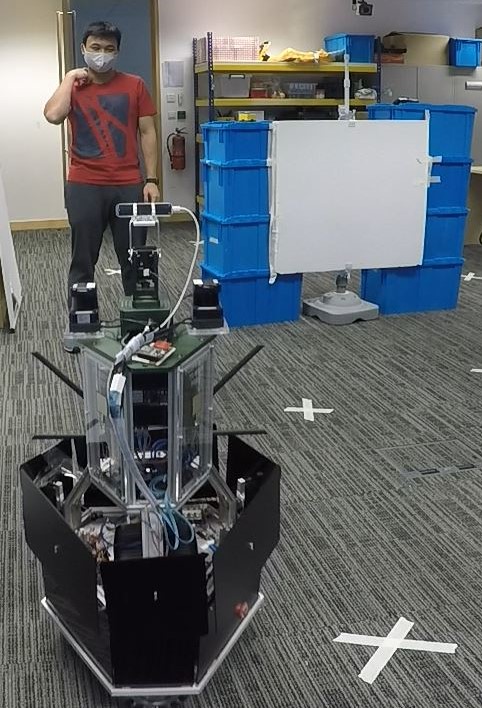} }
\caption{}
\label{fig:give way1}
\end{subfigure}
\begin{subfigure}{0.18\textwidth}
\centerline{\includegraphics[width=0.95\linewidth, height=4cm]{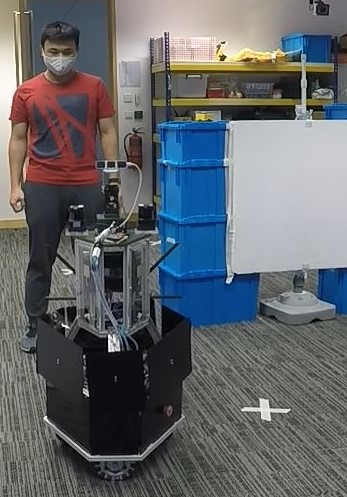}}
\caption{}
\label{fig:give way2}
\end{subfigure}
\begin{subfigure}{0.18\textwidth}
\centerline{\includegraphics[width=0.95\linewidth, height=4cm]{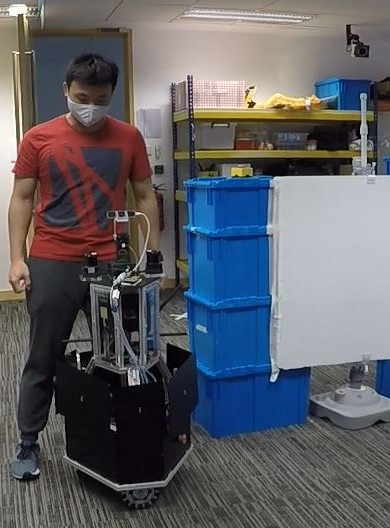}}
\caption{}
\label{fig:give way3}
\end{subfigure}
\begin{subfigure}{0.18\textwidth}
\centerline{\includegraphics[width=0.95\linewidth, height=4cm]{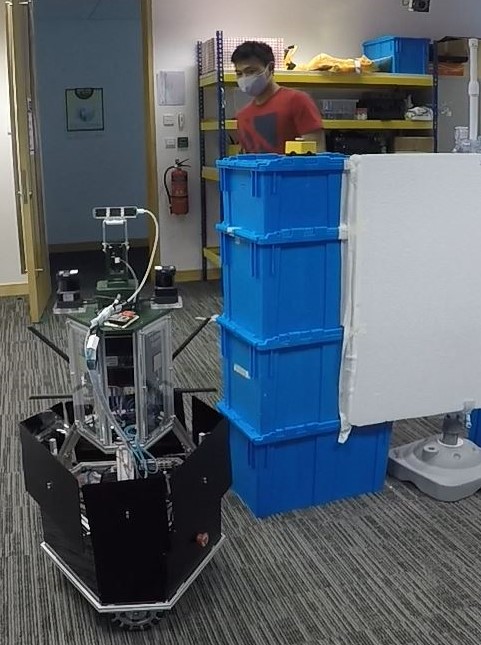}}
\caption{}
\label{fig:give way4}
\end{subfigure}
\begin{subfigure}{0.18\textwidth}
\centerline{\includegraphics[width=0.95\linewidth, height=4cm]{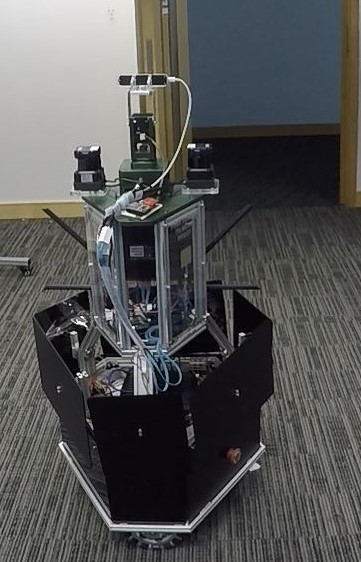}}
\caption{}
\label{fig:give way5}
\end{subfigure}

\begin{subfigure}{0.18\textwidth}
\centerline{\includegraphics[trim=0 0.3cm 0 0, clip, width=0.95\linewidth, height=3cm]{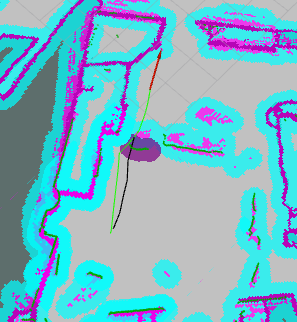} }
\caption{}
\label{fig:give way6}
\end{subfigure}
\begin{subfigure}{0.18\textwidth}
\centerline{\includegraphics[trim=0 0.3cm 0 0, clip, width=0.95\linewidth, height=3cm]{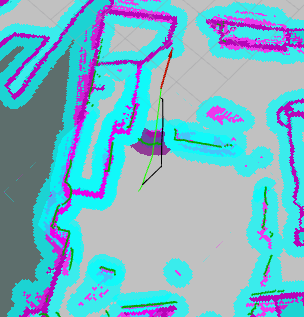}}
\caption{}
\label{fig:give way7}
\end{subfigure}
\begin{subfigure}{0.18\textwidth}
\centerline{\includegraphics[trim=0 0.3cm 0 0, clip, width=0.95\linewidth, height=3cm]{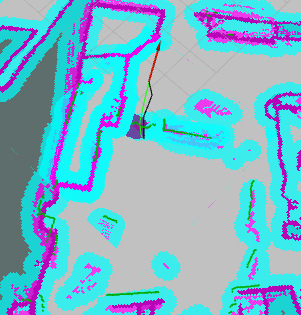}}
\caption{}
\label{fig:give way8}
\end{subfigure}
\begin{subfigure}{0.18\textwidth}
\centerline{\includegraphics[trim=0 0.3cm 0 0, clip, width=0.95\linewidth, height=3cm]{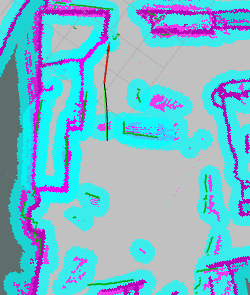}}
\caption{}
\label{fig:give way9}
\end{subfigure}
\begin{subfigure}{0.18\textwidth}
\centerline{\includegraphics[trim=0 0.3cm 0 0, clip, width=0.95\linewidth, height=3cm]{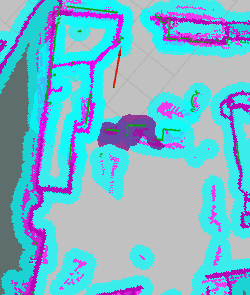}}
\caption{}
\label{fig:give way10}
\end{subfigure}
\caption{Human gave way to the robot after contact. (a)-(e) are the experiment snapshots and (f)-(j) are the corresponding cost map. (a) Initialization of experiment. (b) Robot approach goal via nearest exit. (c) Human unblocked path after contact. (d) Continue to goal via same exit. (e) Goal reached. (f) Path planned to goal. (g) Lower cost (purple) for human before contact. (h) Cost remained the same (purple) after path unblocked. (i) Path planner remained same path. (j) Goal reached.}
\label{fig:give way}
\end{figure*}

The experiment setup is the same in \ref{ASI} as shown in Fig. \ref{fig:give way}. The robot in this case still attempted to reach its goal via the nearer exit on the left, but the path was still shut off in Fig. \ref{fig:give way2}, \ref{fig:give way7}. The robot then attempted to approach human with a gentle touch, mimicking the social cues of natural human behaviour when a path is blocked off in Fig. \ref{fig:give way3}. The proximity filter was kicked into work, repulsed and waited for the timeout. The human cost remained low and permeable before the timeout in Fig. \ref{fig:give way8}. In this scenario, the human decided to unblock the exit in Fig. \ref{fig:give way4}, thus assigned the same lower cost to human in Fig. \ref{fig:give way9}, which allowed the robot continued its path to the goal via the same exit as shown in Fig. \ref{fig:give way5}, \ref{fig:give way10}. This could be potentially useful for time-sensitive or path-sensitive application as it allowed robot to get to the goal as fast as possible without taking the long route while mimicking natural human behaviour to have a gentle ask of unblocking the nearest exit. 

\section{Conclusion}
\label{sec:conclusion}
In this work, we have described the implementation of tactile sensors, proximity filter and Semantic Layer to achieve dynamic human cost assignation, active and passive social interactions with human.
Our proposed method allows the robot to navigation safely and efficiently in crowded situation via delicately designed compliance control strategy which considers both tactile input and laser scan. 
The tactile sensors also endow the robot with the capability of communicating with human more intuitively.
One typical use case is demonstrated in this paper that the robot can gently 'touch' a person who is blocking the exit to check if he is willing to give way, and decide next move subsequently.
With the help of semantic segmentation of vision, different costs can be assigned dynamically to various groups of people within this framework. This promises a socially-aware navigation.
Further work will try to implement human following module for transportation of goods and include human gesture understanding to allow for more intuitive human-robot communication. Besides, 360 camera will be explored to enlarge the view field of robots aiming for more complex and dynamic environments.

\section*{Acknowledgment}
We acknowledge and thank Ng Yung Chuen, Andrew Leong, Lawrence Wong, Liang Wenyu and Chun Ye Tan for their help on experiments and valuable discussion.


\begin{thebibliography}{00}
\bibitem{b1} H. Kretzschmar, M. Spies, C. Sprunk, and W. Burgard, “Socially compliant mobile robot navigation via inverse reinforcement learning,” The International Journal of Robotics Research, Jan. 2016.
\bibitem{b2}R. S. Sutton and A. G. Barto, Reinforcement Learning: An Introduction. Cambridge, MA: Bradford Books, 1998.
\bibitem{b3} Y. F. Chen, M. Everett, M. Liu, and J. P. How, “Socially aware motion planning with deep reinforcement learning,” in 2017 IEEE/RSJ International Conference on Intelligent Robots and Systems, 2017.
\bibitem{b4} C. Chen, Y. Liu, S. Kreiss and A. Alahi, "Crowd-Robot Interaction: Crowd-Aware Robot Navigation With Attention-Based Deep Reinforcement Learning," 2019 International Conference on Robotics and Automation (ICRA), Montreal, QC, Canada, 2019, pp. 6015-6022, doi: 10.1109/ICRA.2019.8794134.
\bibitem{b5} P. Abbeel and A. Y. Ng, “Apprenticeship learning via inverse reinforcement learning,” in Twenty-first international conference on Machine learning - ICML ’04, 2004.
\bibitem{b6} Kretzschmar, Henrik, Markus Spies, Christoph Sprunk, and Wolfram Burgard. “Socially Compliant Mobile Robot Navigation via Inverse Reinforcement Learning.” The International Journal of Robotics Research 35, no. 11 (September 2016): 1289–1307.
\bibitem{b7} D. Fox, W. Burgard and S. Thrun, "The dynamic window approach to collision avoidance," in IEEE Robotics and Automation Magazine, vol. 4, no. 1, pp. 23-33, March 1997
\bibitem{b8} D. Helbing and P. Moln´ar, “Social force model for pedestrian dynamics,” Physical Review E, vol. 51, no. 5, pp. 4282–4286, May 1995.
\bibitem{b9}G. Ferrer and A. Sanfeliu, “Behavior estimation for a complete framework for human motion prediction in crowded environments,” in Proceedings of the 2014 IEEE International Conference on Robotics and Automation (ICRA), May 2014, pp. 5940–5945.
\bibitem{b10} P. Trautman, J. Ma, R. M. Murray, and A. Krause, “Robot navigation in dense human crowds: Statistical models and experimental studies of human–robot cooperation,” The International Journal of Robotics Research, vol. 34, no. 3, pp. 335–356, Mar. 2015.
\bibitem{b11} M. H. Lee and H. R. Nicholls, “Tactile sensing for mechatronics—A state of the art survey,” Mechatronics, vol. 9, no. 1, pp. 1–31, 1999.
\bibitem{b12} T. Matsunaga, K. Totsu, M. Esashi, and Y. Haga, “Tactile display for 2-D and 3-D shape expression using SMA micro actuators,” in Proc. IEEE Annu. Int. Conf. Microtechnologies Med. Biol., 2005, pp. 88–91.
\bibitem{b13} K. Suwanratchatamanee, M. Matsumoto and S. Hashimoto, "Robotic Tactile Sensor System and Applications," in IEEE Transactions on Industrial Electronics, vol. 57, no. 3, pp. 1074-1087, March 2010, doi: 10.1109/TIE.2009.2031195.
\bibitem{b14} D. V. Lu, D. Hershberger, and W. D. Smart, “Layered costmaps for context-sensitive navigation,” in 2014 IEEE/RSJ International Conference on Intelligent Robots and Systems, 2014.
\bibitem{b15} G. J. Paz-Delgado, M. Azkarate, J. R. Sánchez-Ibáñez, C. J. Pérez-del-Pulgar, L. Gerdes and A. J. García-Cerezo, "Improving Autonomous Rover Guidance in Round-Trip Missions Using a Dynamic Cost Map," 2020 IEEE/RSJ International Conference on Intelligent Robots and Systems (IROS), Las Vegas, NV, USA, 2020, pp. 7014-7019.
\bibitem{b16} Fox, Dieter. "Adapting the sample size in particle filters through KLD-sampling." The international Journal of robotics research 22, no. 12 (2003): 985-1003.
\bibitem{b17} Quinlan and O. Khatib, "Elastic bands: connecting path planning and control," Proceedings IEEE International Conference on Robotics and Automation, Atlanta, GA, USA, 1993, pp. 802-807 vol.2.
\end{thebibliography}
\end{document}